\documentclass{article}

\usepackage{amsmath,amssymb}

\newtheorem{myTheo}{Theorem}

\usepackage{times}
\usepackage{graphicx} 

\usepackage{pgfplots,pgfplotstable}

\pgfplotsset{
	jitter/.style={
		y filter/.code={\pgfmathparse{\pgfmathresult+#1}}
	},
	jitter/.default=0.1
}

\usepackage[numbers,sort&compress]{natbib}

\usepackage{algorithm}
\usepackage{algorithmic}

\usepackage{booktabs}
\usepackage{multirow}

\usepackage{rotating}

\allowdisplaybreaks[3]

\title{Latent Regression Bayesian Network \\for Data Representation}

\author{%
	Siqi Nie\footnote{Email: nies@rpi.edu. Affiliation: Rensselaer
		Polytechnic Institute, USA.}\\
	Qiang Ji\footnote{Email: jiq@rpi.edu. Affiliation: Rensselaer Polytechnic Institute, USA.}
}

\date{\today}

\begin{document}

	\maketitle
	
	\begin{abstract} 
	Deep directed generative models have attracted much attention recently due to their expressive representation power and the ability of ancestral sampling. One major difficulty of learning directed models with many latent variables is the intractable inference. To address this problem, most existing algorithms make assumptions to render the latent variables independent of each other, either by designing specific priors, or by approximating the true posterior using a factorized distribution. We believe the correlations among latent variables are crucial for faithful data representation. Driven by this idea, we propose an inference method based on the conditional pseudo-likelihood that preserves the dependencies among the latent variables. For learning, we propose to employ the hard Expectation Maximization (EM) algorithm, which avoids the intractability of the traditional EM by max-out instead of sum-out to compute the data likelihood. Qualitative and quantitative evaluations of our model against state of the art deep models on benchmark datasets demonstrate the effectiveness of the proposed algorithm in data representation and reconstruction.
	\end{abstract} 
	
	\section{Introduction}
	
	Deep directed generative models have received increasing attention recently, because the top-down connections explicitly model the data generating process. Different levels of latent variables capture features (or abstractions \cite{patel2015probabilistic}) in a coarse-to-fine manner. Compared with undirected models such as restricted Boltzmann machines (RBMs) and deep Boltzmann machines (DBMs) \cite{salakhutdinov2009deep}, directed generative models have their own advantages. First, samples can be easily obtained by straightforward ancestral sampling without the need for Markov chain Monte Carlo (MCMC) methods. Second, there is no partition function issue since the joint distribution is obtained by multiplying all local conditional probabilities, which requires no further normalization. Last but most importantly, the latent variables are dependent on each other given the observations through the so-called ``explain-away'' principle. Through their inter dependency, latent variables coordinate with each other to better explain the patterns in the visible layer.
	
	Learning directed models with many latent variables is challenging, mainly due to the intractable computation of the posterior probability. Although Markov Chain Monte Carlo (MCMC) method is a straightforward solution, the mixing stage is often too slow. To simplify the inference for deep belief networks, \citet{hinton2006fast} introduced a complementary prior for the latent variables which makes the posterior fully factorized. Some recent efforts for learning generative models have focused on variational methods \cite{mnih2014neural, kingma2013auto, rezende2014stochastic}, by introducing another distribution to approximate the true posterior and maximize a variational lower bound of the data likelihood. The approximating distribution is typically fully factorized for computational efficiency. However, the assumption of the factorized distribution sacrifices the `` explain-away'' effect for efficient inference, which inevitably enlarges the distance to the true posterior, and weakens the representation power of the model. This defeats a major advantage of directed graphical models.
	
	In this work, we address the problem of learning deep directed models in a different direction. We propose to use the EM algorithm with two approximations in the inference and learning phases. First, we approximate the true posterior distribution during inference by the conditional pseudo-likelihood, which preserves to certain degree the dependencies among latent variables. Second, we approximate the data likelihood using a max-out setting during the E-step of the learning to overcome the exponential number of configurations of the latent variables. As a result, the E-step requires the maximum a posteriori (MAP) inference, which is efficiently solved based on the pseudo-likelihood. It can also be seen as the application of iterated conditional modes (ICM) \cite{besag1986statistical} to directed graphical models. In the M-step, the problem is transferred into parameter learning with complete data, which is much easier to handle.
	
	\section{Related Work}
	
	The research on learning directed model with latent variables can be dated back to 1990s. A standard approach is the Expectation Maximization (EM) algorithm, which maximizes the expected data log-likelihood for parameter learning. EM algorithm and its variants have been used for learning latent mixture of factor analyzers \cite{ghahramani1996algorithm}, probabilistic latent semantic indexing \cite{hofmann1999probabilistic1}, probabilistic latent semantic analysis \cite{hofmann1999probabilistic2} and latent Dirichlet allocation (LDA) \cite{blei2003latent}. Such models have a few latent variables so that the posterior probability of latent variables can be exactly computed in the E-step. 
	
	In the case of many latent variables, exact computation of the posterior is intractable because of the exponential number of the latent variable configurations. \citet{patel2015probabilistic} make one latent variable connecting to a small patch of the input data. Therefore each patch and the corresponding latent variable form a small model, which allows exact maximum a posteriori (MAP) inference. Many other approaches have been proposed to approximate the posterior probability of latent variables along two directions. 
	
	One approach is to replace the true posterior distribution with a factorized distribution as an approximation. This approach was first proposed by \citet{saul1996mean}, known as the mean field theory for learning sigmoid belief networks. A fully factorized variational posterior is introduced to approximate the true posterior distribution of latent variables. Recently, \citet{gan2015learning} extended the mean field method, and proposed a Bayesian approach to learn deep sigmoid belief networks by introducing sparsity-encoraging priors on the model parameters. Alternatively, the posterior distribution can be approximated using a feed-forward network. The wake-sleep algorithm \cite{hinton1995wake} augments the multi-layer belief networks with feed-forward recognition networks. Wake-sleep alternates between updating the model parameters in the wake phase and the recognition network parameters in the sleep phase. Inspired by this idea, many approaches have been proposed recently for learning directed graphical models by maximizing a variational lower bound on the data log-probability. \citet{mnih2014neural} introduced the neural variational inference and learning (NVIL) algorithm for sigmoid belief networks. A feed-forward inference network is used to obtain exact samples from the variational posterior. A neural network is introduced to reduce the variance of the samples. \citet{kingma2013auto} proposed the auto-encoding variational Bayes method for continuous latent variables, in which a reparameterization is employed to efficiently generate samples from the Gaussian distribution. Similarly, \citet{rezende2014stochastic} propose a stochastic backpropagation algorithm for learning deep generative models with continuous latent variables. \citet{gregor2013deep} augment the directed model with an encoder, which is also kind of inference network.
	
	Another direction is to make the posterior probability factorized by specifically designing a prior distribution of latent variables. \citet{hinton2006fast} proposed a complementary prior to ensure a factorized posterior, and proposed a fast learning algorithm for deep belief networks (DBNs), which is basically a hybrid network with a single undirected layer and several directed layers.
	
	In all the above-mentioned methods, the inference typically assumes independency among latent variables due to special prior or factorized approximation in order to accelerate the inference. Because of this assumption, the inference network is not able to capture the correlations among the latent variables. Therefore the approximate posterior might differ significantly from the underlying true posterior. In this work, we intend to preserve the latent variable dependencies for better data representation. We approximate the posterior probability by the conditional pseudo-likelihood, and employ a hard version of the EM algorithm for parameter estimation.
	
	
	\section{Latent Regression Bayesian Network}
	We propose a generalized directed graphical model, called latent regression Bayesian network (LRBN), as shown in Fig. \ref{fig:model} (a). The latent variables in LRBN are binary, and the visible variables can be continuous or discrete. Each latent variable is connected to all visible variables. We discuss the parameterization of both cases in the sequel. The case of continuous latent variables can be referred to as factor analyzers \cite{ghahramani1996algorithm, tang2012deep} or deep latent Gaussian models \cite{kingma2013auto, rezende2014stochastic}.
	
	\subsection{Discrete LRBN}
	
	For discrete LRBN, both latent and observation variables are discrete. For brevity, we discuss the binary case with observation variables $x\in \mathbb{B}^{n_d}$ and latent variables $h\in \mathbb{B}^{n_h}$. We assume that the latent variables $h$ determine the patterns in data $x$, therefore directed links are used to model their relationships, as in a Bayesian network.
	
	Prior probability for latent variables is represented as a log-liner model,
	\begin{equation}
		P(h_j=1) = \sigma(d_j)\, ,
		\label{prior}
	\end{equation}
	where $d_j$ is the parameter defining the prior distribution for node $h_j$; $\sigma(\cdot)$ is the sigmoid function $\sigma(z)=1/(1+e^{-z})$. 
	
	The conditional probability given the latent variables is,
	\begin{equation}
		P(x_i=1|h)=\sigma(\sum_j w_{ij} h_j+b_i)\, ,
		\label{condi}
	\end{equation}
	where $w_{ij}$ is the weight of the link connecting node $h_j$ and $x_i$; $b_i$ is the offset for node $x_i$. The joint probability is,
	\begin{equation}
		P_\theta(x,h)\!=\!\frac{\exp\!\left(\sum_{i,j}\! w_{ij} x_i h_j\!+\!\sum_i\! b_i x_i\!+\!\sum_j\! d_j h_j\! - \sum_i \log \left(1+\exp(\sum_jw_{ij}h_j+b_i)\right)\right)}{\prod_j\!\left(\!1\!+\!\exp(d_j)\!\right)}\, .
		\label{BNjointBinary}
	\end{equation}
	In this case, the model becomes a sigmoid belief network (SBN) \cite{neal1992connectionist} with one latent layer. If more layers are added on top, the conditional probability is defined in the same way as Eq. \ref{condi}. The model is named a regression model based on the nature of the conditional probability. For discrete visible node, the input to the sigmoid function is a linear combination of the latent variables; for continuous visible node, the mean of a visible node is a linear combination of its parent nodes.
	
	\begin{figure}[b]
		\centering      
		\includegraphics[width=0.8\textwidth]{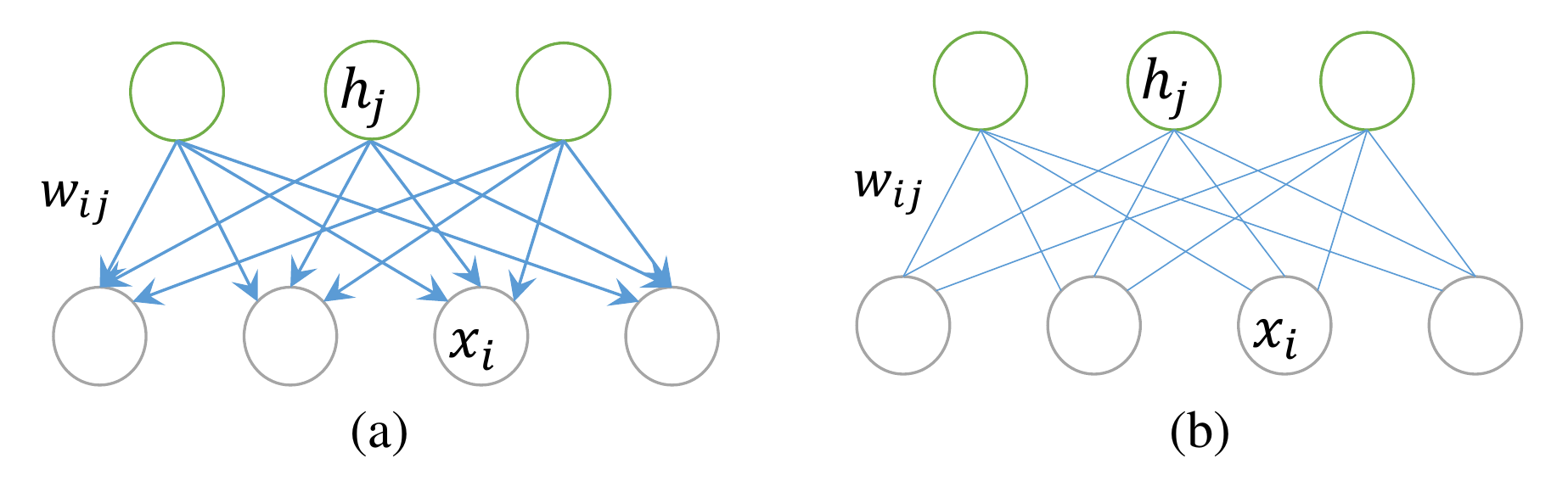}
		\caption{Graph representation of the (a) directed and (b) undirected model for data representation. Each link is associated with a weight parameter.}
		\label{fig:model}
	\end{figure}

	As a similar model with undirected links, RBMs (Fig. \ref{fig:model} (b)) have been widely used in the literature for feature learning and data representation. The joint probability defined by a discrete RBM is,
	\begin{equation}
		P_{\text{RBM}}(x,h)\!=\!\frac{1}{Z} \exp\! \left(\!\sum_i b_i x_i \!+\! \sum_{i,j}w_{ij}x_i h_j\!+\!\sum_j d_j h_j\!\right)\, .
		\label{RBMjointBinary}
	\end{equation}
	Comparing Eq. \ref{BNjointBinary} and \ref{RBMjointBinary}, the additional terms in the numerator captures the correlations among them. This is the reason why $P(h|x)$ is not factorized over individual latent nodes $h_j$ in LRBN, which is the major difference from RBM. An advantage of Eq. \ref{BNjointBinary} is that every term can be computed given the values of all variables, without the issue of the intractable partition function $Z$.
	
	The discussion of hybrid LRBN is moved to a supplementary material due to the limited space.
	
	\subsection{Hybrid LRBN}
	
	For hybrid LRBN, the observation variables $x\in \mathbb{R}^{n_d}$ are continuous while the latent variables are binary $h\in \mathbb{B}^{n_h}$. The prior distribution of the latent variables is the same as Eq. \ref{prior}. Given the latent variables, the visible variable is assumed to follow Gaussian distribution, whose mean is a linear combination of the latent variables,
	\begin{equation}
		P\left(x_i|h\right)\sim\mathcal{N}\left(\sum_j w_{ij}h_j +b_i, \, \sigma_i\right),\quad i=1,\dots,n_d\, ,
		\label{condiprob}
	\end{equation}
	where $w_{ij}$ is the weight of the link connecting node $h_j$ and $x_i$; $b_i$ is the offset of the mean for node $x_i$; $\sigma_i$ is the standard deviation. To simplify the learning process, each component of the data is normalized to have zero mean and unit variance, therefore $\sigma_i$ is set to 1. From the prior distribution and conditional distribution, the joint distribution for $x$ and $h$ is,
	\begin{equation}
		P_\theta(x,h)=\frac{\exp\left(-\frac{1}{2}||x-Wh-b||^2 + d^Th\right)}{(2\pi)^{n_d/2}\prod_j\left(1+\exp(d_j)\right)}\, ,
		\label{joint}
	\end{equation}
	or,
	\begin{equation}
		\begin{split}
			& P_\theta(x,h)=\frac{1}{(2\pi)^{n_d/2}\prod_j\left(1+\exp(d_j)\right)} \\
			& \exp\!\left(\frac{1}{2}(x\!-\!b)^T(x\!-\!b)\!-\!x^TWh\!+\!\frac{1}{2}h^TW^TWh\!-\!d^Th\right)\, .
		\end{split}
		\label{jointRBN}
	\end{equation}
	For brevity, vector and matrix forms are used, $W=\{w_{ij}\}$, $b=\{b_i\}$, $d=\{d_j\}$. $\theta=\{W,b,d\}$ represent all the parameters.
	
	For real-valued input data and binary latent variables, Gaussian-Bernoulli RBM defines the joint probability of visible and latent layer,
	\begin{equation}
		P_{\text{RBM}}(x,h)\!\!=\!\!\frac{1}{Z} \exp\! \left(\!-\frac{1}{2}(x\!-\!b)^T(x\!-\!b) \!+\! x^TWh\!+\!d^Th\!\right)\, ,
		\label{jointRBM}
	\end{equation}
	where $Z$ is the partition function to make $P_{\text{RBM}}(x,h)$ a valid probability distribution.
	The input data is assumed to be normalized to have unit variance.
	
	Comparing LRBN (Eq. \ref{jointRBN}) and RBM (Eq. \ref{jointRBM}), with the same dimensionality of the visible and latent layer, the two models have the same amount of parameters. However, the directed model has a quadratic term $h^TW^TWh=\sum_i \left(\sum_j w_{ij}h_j\right)^2$, which does not exist in the joint distribution of RBM. This term explicitly captures the correlations among the latent variables $h$. It also explains why given the visible layer, the latent variables are dependent on each other.

	\section{LRBN Inference}
	\label{sec:4}
	In this section, we introduce an efficient inference method for LRBN based on conditional pseudo-likelihood. Given a LRBN model with known parameters, the goal of inference is to compute the posterior probability of the latent variables given input data, i.e., computing $P(h|x)$. 
	
	In this work, we are interested in the maximum a posteriori (MAP) inference, which is to find the configuration of latent variables that maximizes the posterior probability given observations,
	\begin{equation}
		h^*=\arg\max_h P(h|x)\, .
	\end{equation}
	The MAP inference is motivated by the observation that from the data generating point of view, the variables in one latent layer take values according to the conditional probability given its upper layer. Therefore this configuration dominates all the others in explaining each data sample. In addition, the goal of feature learning is to learn a feature $h$ that best explains $x$. In this regard, we only care about the most probable states of the latent variables given the observation.
	
	Because of the dependencies among elements of $h$, direct computing $P(h|x)$ is computationally intractable, in particular when the dimension of $h$ is high. According to the chain rule, the posterior probability of $h$ is, 
	\begin{equation}
		P(h|x)=\prod_j P(h_j|h_1,\dots,h_{j-1},x)\, ,
	\end{equation}
	The pseudo-likelihood replaces the conditional likelihood by a more tractable objective, i.e.,
	\begin{equation}
		P(h|x)\approx \prod_j P(h_j|h_{-j},x)\, ,
	\end{equation}
	where $h_{-j}=\{h_1,\dots,h_{j-1},h_{j+1},\dots,h_{n_h}\}$ is the set of all latent variables except $h_j$. In this approximation, we add conditioning over additional variables.
	
	The conditional pseudo-likelihood can be factorized into local conditional probabilities, which can be estimated in parallel. To optimize over the pseudo-likelihood, one latent variable is updated by fixing all other variables,
	\begin{equation}
		h_j^{t+1}=\arg\max_{h_j} P(h_j|x, h^t_{-j})\, , \quad 1\leq j \leq n_h\, ,
		\label{update}
	\end{equation}
	where $t$ denotes the $t^{\text{th}}$ iteration.
	\begin{myTheo}
		The updating rule (Eq. \ref{update}) guarantees that the posterior probability $P(h|x)$ will only increase or stay the same after each iteration.
		\begin{equation}
			P(h_j^{t+1},h^t_{-j}|x)\geq P(h^t|x)\, .
		\end{equation}
	\end{myTheo}
	
	The conditional probability of one latent variable by fixing all other variables is easy to compute, since it involves the computation of the joint distribution twice.
	\begin{equation}
		P(h_j|x,h_{-j})\!=\!\frac{P(h_j,x,h_{-j})}{\sum_{h'_j}P(h'_j,x,h_{-j})}\, .
	\end{equation}
	In general, computing the joint probability $P(x,h)$ has complexity $O(n_d n_h)$. If each latent variable is updated $t$ times to get $h^*$, the overall complexity for the LRBN inference is $O(tn_dn_h^2)$, which is much lower than the $O(2^{n_h}n_dn_h)$ complexity when computing $P(h|x)$ directly.
	
	The updating method can be seen as a coordinate ascent algorithm or the iterated conditional modes (ICM) as in inference of Markov random fields. Typically in MRF the number of neighbors for one node is limited. In the case of LRBN, one latent variable is related to all the other variables, due to the rich dependencies encoded in the structure. As discussed above, existing methods to address the inference intractability problem makes the posterior probability completely factorized, therefore sacrificing the dependencies among the latent variables. In contrast, through pseudo-likelihood, we can preserve the dependencies to certain degree.
	
	The inference method requires an initialization for the hidden variables. Different initializations will end up with different local optimal points. To obtain consistent initialization, we drop the direction of the links, and treat the directed model as an undirected one. Therefore, the latent variables are independent of each other given the observations. Specifically, for binary input, 
	\begin{equation}
		P(h_j=1|x) = \sigma(\sum_i w_{ij}x_i+d_j)\, .
		\label{eqn:ini1}
	\end{equation}
	For continuous input, based on Eq. \ref{jointRBN}, we drop the off-diagonal terms of matrix $W^TW$ for the sake of efficiency, resulting in a factorized distribution of the latent variables,
	\begin{equation}
		P(h_j=1|x) = \sigma(\sum_i w_{ij}x_i + d_j -s_j)\, ,
		\label{eqn:ini2}
	\end{equation}
	where $(s_1,\dots, s_{n_h}) = diag(\frac{1}{2}W^TW)$. For the initialization, the dependency among the latent variables is ignored, and then through coordinate ascent, it is recovered by updating a subset of variables with others fixed.
	
	\section {LRBN Learning}
	
	In this section, we introduce an efficient LRBN learning method based on the hard Expectation Maximization (EM) algorithm. The conventional EM algorithm is not an option here due to the intractability of computing posterior probability in the E-step. The hard version of EM algorithm has been explored in \cite{van2014factoring} for learning a deep Gaussian mixture model. This model has a deep structure in terms of linear transformations, but only has two layers of variables. 
	
	\subsection{Learning One Latent Layer}
	
	Consider the model $P_\theta(x,h)$ defined in section 3. The goal of parameter learning is to estimate the parameters $\theta=\{w,b,d\}$ given a set of data samples $D\!=\!\{x^{(m)}\}_{m=1}^M$. The conventional maximum likelihood (ML) parameter estimation is to maximize the following objective function,
	\begin{equation}
		\theta^*=\arg\max_\theta \sum_m \log \sum_h P_\theta(x^{(m)},h)\, .
		\label{MLobj}
	\end{equation} 
	The second summation in Eq. \ref{MLobj} is intractable due to the exponentially many configurations of $h$. In this work, we employ a max-out estimation of the data log-probability, with the following objective function, 
	\begin{equation}
		\theta^* = \arg \max_\theta \sum_m \log \max_h P_\theta (x^{(m)},h)\, .
		\label{objhardEM}
	\end{equation}
	
	Note that the max-out approximation of the data likelihood is not equivalent to approximating $P(h|x)$ with a delta function. A delta function must be avoided since it is also factorized, which defeats our motivation of preserving the dependency.
	
	With objective function Eq. \ref{objhardEM}, the learning method becomes a hard version of the EM algorithm, which iteratively fills in the latent variables and update the parameters. In the E-step,  $h^*=\arg\max_h P(x,h)$ is effectively estimated using the proposed inference method. In the M-step, the problem of parameter estimation is straightforward because now we are dealing with complete data. 
	
	For binary observations, the parameter learning can be decomposed into learning local CPD for each variable, by solving multiple logistic regression problems. The gradient of the parameters is,
	\begin{equation}
		\frac{\partial \log P(x,h)}{\partial w_{ij}}= h_j \left(x_i - P(x_i=1|h)\right)\, ,
		\label{eqn:gradient1}
	\end{equation}
	\begin{equation}
		\frac{\partial \log P(x,h)}{\partial b_i}= x_i - P(x_i=1|h)\, .
		\label{eqn:gradient2}
	\end{equation}
	In hybrid LRBN, the objective function is convex, and the maximization likelihood solution can be obtained by setting,
	\begin{equation}
		\sum_m \frac{\partial \log P(x^{(m)},h^{(m)})}{\partial \theta}=0\, .
	\end{equation}
	The solution of parameters has a closed form,
	\begin{equation}
		W = \left(\sum_m (x^{(m)}-b)(h^{(m)})^T\right)\left(\sum_m h^{(m)}(h^{(m)})^T\right)^{-1}\, .
	\end{equation}
	In case of large datasets, it is time consuming because all the training instances are used to compute the gradient. Stochastic gradient ascent algorithm can be used to address this issue. The true gradient is approximated by the gradient at a minibatch of training samples. As the algorithm sweeps through the entire training set, it performs the gradient updating for each training sample. Several passes is made over the training set until the algorithm converges. The learning method is summarized in Algorithm \ref{algo:lrbn}.
	
	\begin{algorithm}[t]
		\caption{Unsupervised Parameter learning of an LRBN with one latent layer.}
		\label{algo:lrbn}
		\begin{algorithmic}[1]
			\item[\textbf{Input}] training data $X=\{x^{(m)}\}$
			\item[\textbf{Output}] parameters $\theta$ of an LRBN
			\STATE Initial parameters $\theta$;
			\STATE Initialize the states $h_0$ for all the latent variables using some feed-forward model (Section \ref{sec:4});	
			\WHILE{parameters not converging, }
			\STATE Random select a minibatch of data instances $x \in X$
			\STATE Update the corresponding $h$ for $x$ by maximizing the posterior probability, using current parameters,
			\begin{equation}
				h^* = \arg \max_h P_\theta(x,h)\, .
			\end{equation}
			\STATE Compute the gradients using Eq. \ref{eqn:gradient1} and \ref{eqn:gradient2}. Update the parameters,
			\begin{equation}
				\theta = \theta + \lambda \triangledown_\theta \log P(x,h^*)
			\end{equation}
			\ENDWHILE
		\end{algorithmic}
	\end{algorithm}

	\subsection{Learning Deep Layers}
	\label{sec5.2}
	
	The learning method of a two-layer LRBN not only provides the parameter $\theta$, but also perform the MAP inference to obtain $h^*\!=\!\arg\max_h P(h|x)$. If we denote the features as $h^1$ and treat them as the input to another LRBN, the same learning procedure can be repeated to learn another layer of features $h^2$. By doing this we stack another LRBN on top of the first one to build a deep model. 
	
	In general, let $h^{l}$ denote the variables in the $l^{\text{th}}$ latent layer ($0\leq l\leq L, h^{0}\!=\!x$), and $\theta^l$ be the parameters involved between layer $l$ and $l\!+\!1$. 
	
	The parameter $\theta^{l*}$ is estimated as,
	\begin{equation}
		\theta^{l*} = \arg \max_{\theta^l} \sum_m \log \max_{h^{l+1}} P_\theta (h^{l,(m)},h^{l+1})\, , \quad 1\leq l \leq L\, .
		\label{eqn:param}
	\end{equation}
	To optimize the objective function, we use the stochastic gradient ascent method, and replace the input $X$ by $h^l$ in Algorithm \ref{algo:lrbn}. By performing the layer-wise learning procedure from the first latent layer to the $L^{\text{th}}$, we learn a deep model from bottom-up sequentially. Each time the MAP estimation of one latent layer is treated as input to the next two-layer model. For data instance $x^{(m)}$, 
	\begin{equation}
		h^{l,(m)}=\arg\max_{h^l}P(h^l|h^{l-1,(m)})\, , \quad 1\leq l \leq L\, ,
	\end{equation}
	where $h^{0,(m)}=x^{(m)}$. 
	The layer-wise pre-training procedure extracts different levels of features from the input data, and also provides an initial estimation of the parameters.
	
	\subsection{Fine Tuning}
	The layer-wise training ignores the parameters of other layers when training a model for each layer from bottom-up. To improve the model performance globally, we employ a fine tuning procedure from top-down after the layer-wise pre-training phase. Depending on whether the labels are available or not, the fine tuning can be done in either supervised or unsupervised manner as discussed below.
	
	\subsubsection{Unsupervised Fine Tuning}
	\label{sec5.3.1}
	
	By extending Eq. \ref{objhardEM}, the objective function for learning with multiple latent layers is,
	\begin{equation}
		\theta^* = \arg \max_\theta \sum_m \log \max_{\{h^{l}\}} P_\theta (x^{(m)},h^{1},\dots,h^{L})\, .
		\label{objhardEMdeep}
	\end{equation}
	Given the states in layer $l$, the variables in layer $l-1$ is independent of the variables in layer $l+1$. Therefore, the unsupervised fine-tuning is performed for every three consecutive layers. The variables in the middle layer is updated with its upper and lower layers fixed,
	\begin{equation}
		h^{l*} = \arg \max_{h^{l}}P(h^{l}|h^{l-1},h^{l+1})\, , \quad 1\leq l \leq L-1\, .
		\label{eqn:map}
	\end{equation}
	The conditional probability $P(h^{l}|h^{l-1},h^{l+1})$ is also approximated by the conditional pseudo-likelihood,
	\begin{equation}
		P(h^{l}|h^{l-1},h^{l+1})\approx \prod_j P(h_j^l|h_{-j}^{l},h^{l-1},h^{l+1})\, .
	\end{equation}
	MAP inference is performed by updating one variable with all others fixed,
	\begin{equation}
		h_j^{t+1}=\arg\max_{h_j} P(h_j^l|h_{-j}^{l},h^{l-1},h^{l+1})\, , \quad 1\leq j \leq n_h\, .
	\end{equation}
	To be consistent with bottom-up training, the initialization of the latent variable $h^l$ always follows Eq. \ref{eqn:ini1} and \ref{eqn:ini2}.
	
	With the updated layer $h^{l}$, we are able to update the parameters $\theta^{l-1}$ and $\theta^l$ through,
	\begin{equation}
		\theta^{l-1*}=\arg\max_\theta \sum_m \log P(h^{l-1,(m)},h^{l,(m)})\, .
		\label{param1}
	\end{equation}
	and
	\begin{equation}
		\theta^{l*}=\arg\max_\theta \sum_m \log P(h^{l,(m)},h^{l+1,(m)})\, .
		\label{param2}
	\end{equation}
	This process alternates between parameters updating and latent states updating. Therefore, the information is able to propagate among different layers, and the overall quality of the model will increase. The fine tuning proceeds in a top-down manner starting from layers $\{L\!-\!2,L\!-\!1,L\}$ and ending with layers $\{0,1,2\}$. The bottom-up pre-training and top-down fine-tuning procedures are performed iteratively until the parameters converge.
	
	\subsubsection{Supervised Fine Tuning}
	
	The parameter of the model can be fine-tuned discriminatively if the label information is available. Define a set of target variables $t\!=\!(t_1,...,t_C)$, with $C$ being the total number of classes. $t_c=1$ if a sample belongs to class $c$, and $t_k\!=\!0, \forall k\!\neq\! c$. The supervised fine-tuning contains three steps. First, a layer-wise pre-training is performed with $L\!-\!1$ latent layers (excluding the top layer) using the method discussed in Section \ref{sec5.2}, obtaining $\theta^l$ and $h^l, 1\!\leq\! l\! \leq\! L\!-\!1$. 
	
	Second, the parameter $\theta^L$ for the top two layers is estimated as, 
	\begin{equation}
		\theta^{L-1*}=\arg\max_\theta \sum_m \log P(h^{L-1,(m)},t^{(m)})\, .
	\end{equation}
	This step works with complete data, which does not require inference because $t$ is always observed. Thees two steps form the bottom-up pre-training procedure. 
	
	Third, the top-down fine tuning starts with layers $\{L\!-\!2,L\!-\!1,L\}$ with $h^L=t$ and ends with layers $\{0,1,2\}$. Latent states updating follows Eq. \ref{eqn:map}, and parameter updating follows Eq. \ref{param1} and \ref{param2}.
	
	\section{Experiments}
	In this section, we evaluate the performance of LRBN and compare against other methods on three binary datasets: MNIST, Caltech 101 Silhouettes and OCR letters. Binary datasets are chosen to compare with other models. The extension to real-value datasets is straightforward. The experiments will evaluate representation and reconstruction power of the proposed model.

	\subsection{Experimental protocol}

	We trained the LRBN model using stochastic gradient ascent algorithm with learning rate 0.25. The size of the minibatches is set to 20. Two different structures are studied: one hidden layer with 200 variables, and two hidden layers with each layer containing 200 variables, consistent with the configurations in \cite{mnih2014neural, gan2015learning}. For each dataset, we randomly selected 100 samples from the training set to form a validation set. The joint probability on the validation set is  a criterion for early stopping.
	
	In this section, we first evaluate the MAP configuration of the latent variables through reconstruction. Reconstruction is performed as follows: given a data vector $x$, perform a MAP inference to get $h^*\!=\!\arg\max_h P(h|x)$. Then $\tilde{x}\!=\!\arg\max_x P(x|h^*)$ is the reconstructed data. The reconstruction error $|\tilde{x}-x|^2$ can evaluate how well the model fits the data.
	
	The second criterion is the widely used test data log-probability. Directly computing probability $P(x)$ is intractable due to the exponentially many terms in the summation $P(x)\!=\!\sum_h P(x,h)$. In this work, we estimate the log-probability using the conservative sampling-based log-likelihood (CSL) method \cite{bengio2013bounding},

	\begin{equation}
	\log \hat{P}(x)=\log \text{mean}_{h\in S} P(x|h)\, ,
	\vspace{-3pt}
	\end{equation}
	where $S$ is a set of samples $h$ of the latent variables collected from a Markov chain. The expectation of the estimator is a lower bound on the true log-likelihood \cite{bengio2013bounding}. Because of the nature of directed models, samples can be collected from the ancestral sampling procedure. Specifically, the top layer is sampled from the prior probability $P(h^L)$, and the lower layers are sampled from the conditional probabilities $P(h^{l-1}|h^l)$. One million samples are used to reach convergence of the estimation of log-probability, and the average of ten repetitions is reported. 
	
	The reconstruction error evaluates the quality of the most probable explanation of the latent variables given observations, while the data log-probability evaluates the overall quality of all configurations of latent variables. They are two complementary criteria for model evaluation. 
	
	In the experiment, we compare with published results if they are available. For reconstruction we implement the NVIL, RBM, DBN and DBM models following \cite{mnih2014neural,cho2013enhanced, hinton2006fast,salakhutdinov2009deep} (denoted by (*) in the following tables). Similar log-likelihood achieved by our implementation indicates the correctness of the implementation.

	\subsection{MNIST dataset}

	The first experiment is performed on the binary version of the MNIST dataset (thresholding at 0.5). The dataset consists of 70,000 handwritten digits with dimension $28\!\times\! 28$. It is partitioned into a training set of 60,000 images and a testing set of 10,000 images. 
	
\begin{figure}[ht]
			\centering
			\includegraphics[width=0.6\textwidth]{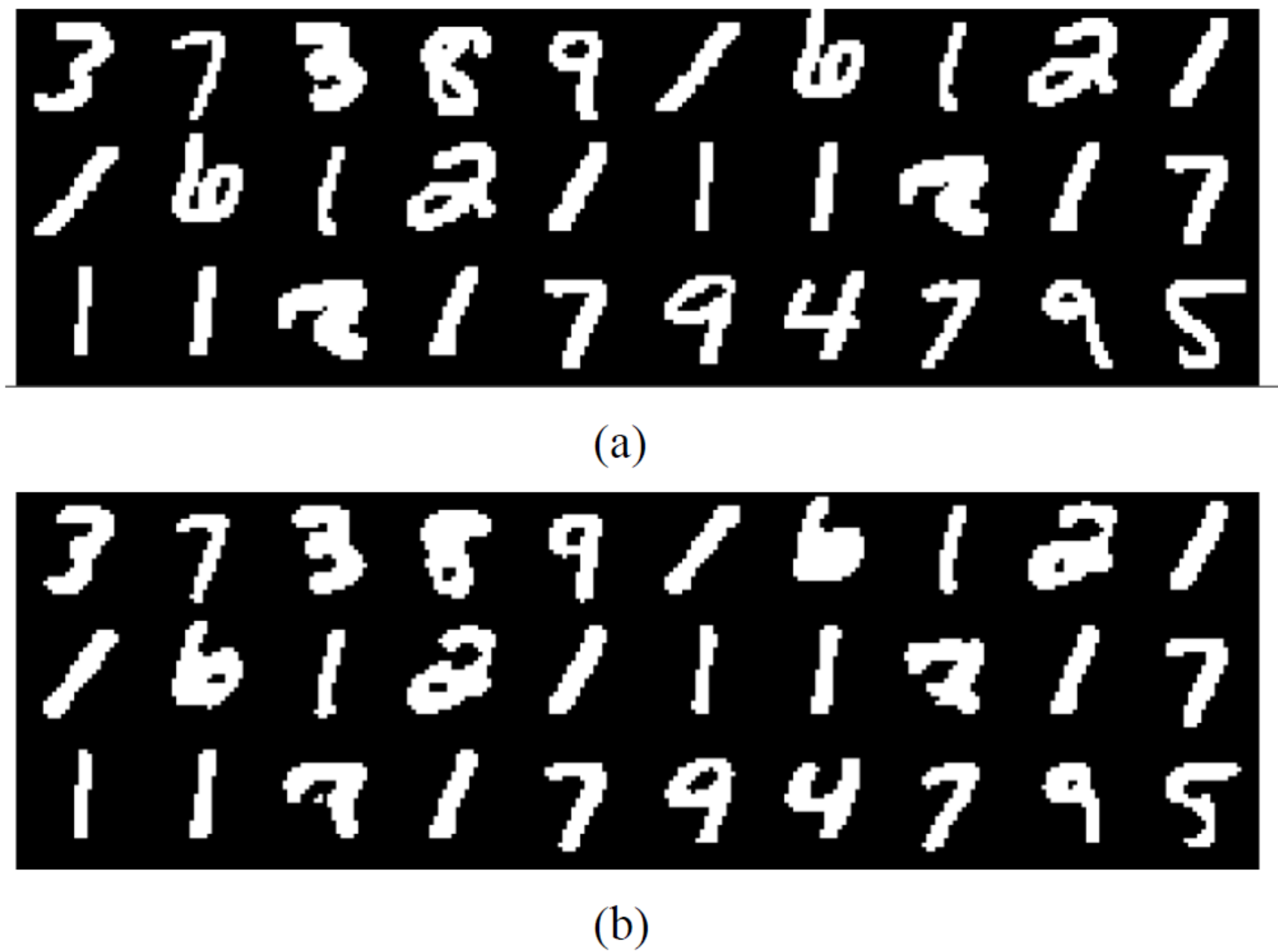}       
			\caption{Examples of the reconstruction. (a) Original digit images. (b) reconstructed by the proposed model.}
			\label{fig:recon}
		\end{figure}

	The average reconstruction errors of different learning models are reported in Table \ref{reconmnist}. The MAP inference of neural variational inference and learning (NVIL) \cite{mnih2014neural} is through the inference network. For deep belief network (DBN) \cite{salakhutdinov2008quantitative} and deep Boltzmann machine (DBM) \cite{salakhutdinov2009deep}, the posterior $P(h|x)$ is already factorized, so that the inference is performed individually for each latent variable. The average reconstruction error of the proposed model is 4.56 pixels, which significantly outperforms the other competing methods. This is consistent with our objective function, indicating the most probable explanation contains most information in the input data, which is effectively captured in the proposed model. Some examples of the reconstruction are shown in Fig. \ref{fig:recon}.

	\begin{table}[ht]
		\centering
		\small
		\caption{Average reconstruction errors of different methods on MNIST dataset. (*) represents our own implementation, same for the following tables.}
		\begin{tabular}{l|l|c}
			\hline
			Method & DIM & Recon Error \\
			\hline
			NVIL* \cite{mnih2014neural} &200 - 200 & 35.52 \\
			DBN* \cite{salakhutdinov2008quantitative} & 200 - 200 & 29.78 \\
			DBM* \cite{salakhutdinov2009deep} & 200 - 200 & 23.52 \\\hline
			LRBN & 200 - 200 & 4.56 \\\hline
		\end{tabular}
		\label{reconmnist}
	\end{table}

	\begin{table}[ht]
		\centering
		\small
		\caption{Test data log-probabilities of different models using on MNIST dataset.}
		\begin{tabular}{l|l|c}
			\hline
			Method & DIM & 10k \\
			\hline
			VB \cite{gan2015learning} & 200 & -116.91 \\
			VB \cite{gan2015learning} & 200 - 200 & -110.74 \\
			NVIL \cite{mnih2014neural} & 200 & -113.1 \\	
			NVIL \cite{mnih2014neural} & 200 - 200 & -99.8 \\	
			DBN \cite{salakhutdinov2008quantitative} & 500 - 2000 & -86.22 \\
			DBM \cite{salakhutdinov2009deep} & 500 - 1000 & -84.62 \\
			\hline
			LRBN & 200 & -108.7\\
			LRBN & 200 - 200 &  -100.3\\
			\hline 
		\end{tabular}
		\label{mnistloglik}
	\end{table}
	
		\begin{figure}[ht]
			\centering
			\includegraphics[width=0.6\textwidth]{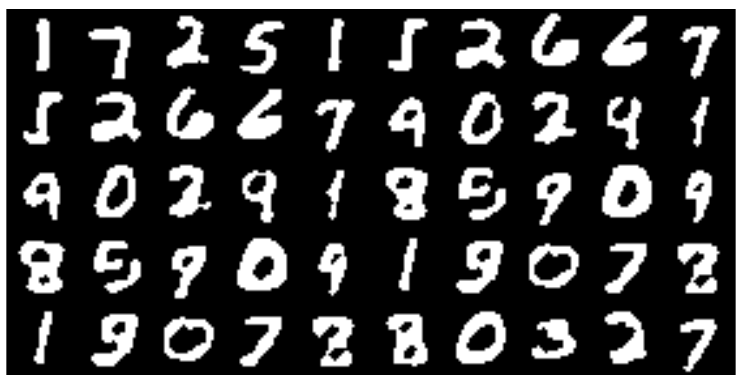}       
			\caption{Random samples from the generative model on MNIST dataset.}
			\label{fig:digits}
		\end{figure}

	In Table \ref{mnistloglik} we report the average log-probability on the test set. With the same dimensionality, LRBN outperforms variational Bayes \cite{gan2015learning}, and is similar to that learned using NVIL \cite{mnih2014neural}. Even though our objective function does not explicitly maximize the data likelihood, the learned model achieves comparable performance compared with state of the art learning methods, which indicates that the proposed method is also effective in capturing the distribution of the training data. In all algorithms, introducing a second hidden layer improves the performance. Our method achieves almost 8 nats improvement when additional latent layer is used. Some samples from the generative model are given in Fig. \ref{fig:digits}. 
	
	There is still a gap between the proposed learning method and DBN or DBM. One reason is that we do not have as many latent nodes as in DBN and DBM. Moreover, the max-out approximation of the data likelihood during learning drops all the non-dominant configurations of the latent variables. Therefore it does not necessarily perform well on the task of likelihood comparison. However, it still achieves comparable or even better performance compared to other learning methods.

	\subsection{Caltech 101 Silhouettes dataset}

	The second experiment is performed on the Caltech 101 Silhouettes dataset. The dataset contains 6364 training images and 2307 testing images. The reconstruction error is reported in Table \ref{reconsil}. The proposed learning method outperforms all the competing methods by a large margin, indicating the effectiveness of the max-out approximation.
	
\begin{figure}[ht]
		\centering
		\includegraphics[width=0.6\textwidth]{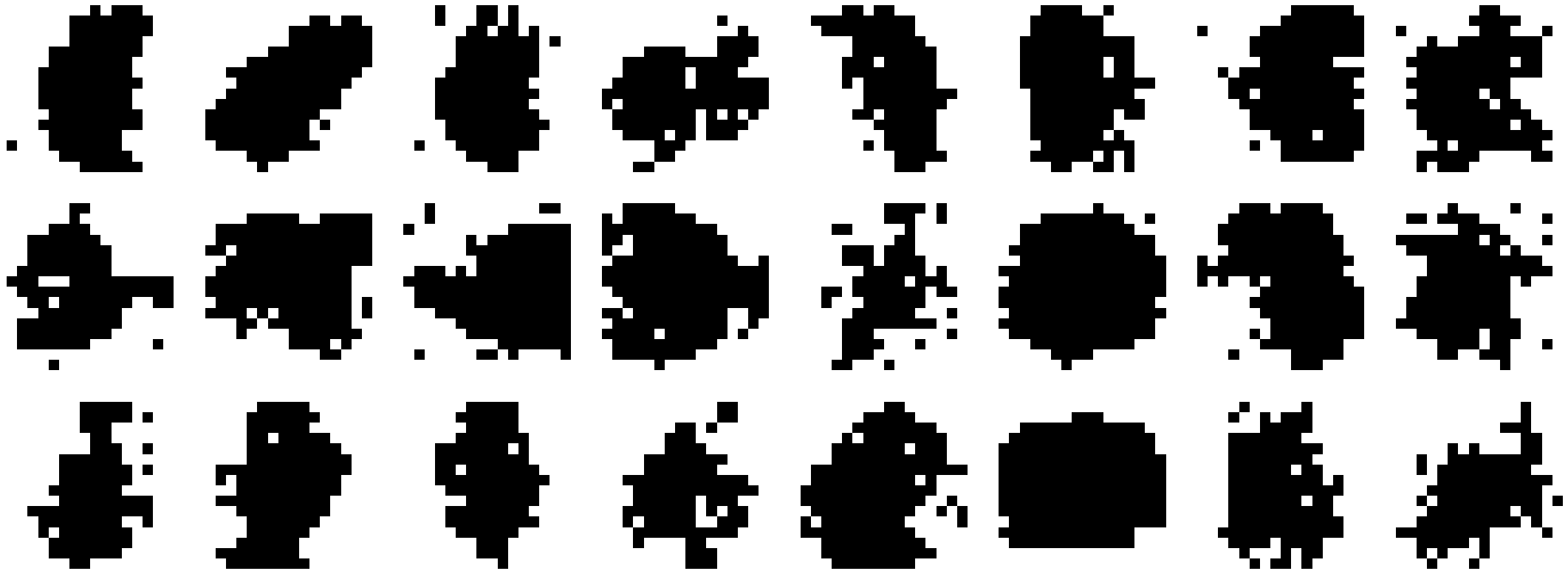}       
		\caption{Random samples from the generative model on Caltech 101 Silhouettes dataset.}
		\label{fig:sil}
	\end{figure}

	The test data log-probability is reported in Table \ref{silloglik}. With the same dimensionality, the model learned by the proposed algorithm outperforms the one learned by variational Bayes \cite{gan2015learning}, which is considered as one of the state of the art methods of training sigmoid belief networks. With one hidden layer of size 200, the improvement is 34 nats; with a second hidden layer of size 200, the improvement is 28 nats. Moreover, compared to an RBM with much more parameters, our model also achieves better performance, indicating the importance of the underlying dependency of the latent variables. Examples are shown in Fig. \ref{fig:sil}.
	
	\begin{table}[ht]
		\centering
		\small
		\caption{Average reconstruction errors of different methods on Caltech 101 Silhouettes dataset.}
		\begin{tabular}{l|l|c}
			\hline
			Method & DIM & Recon Error \\
			\hline
			NVIL* \cite{mnih2014neural} &200 - 200 & 29.78 \\
			RBM* \cite{cho2013enhanced} & 200 & 32.47\\
			DBN* \cite{salakhutdinov2008quantitative} & 200 - 200 & 28.17 \\
			DBM* \cite{salakhutdinov2009deep} & 200 - 200 & 24.90 \\\hline
			LRBN & 200 - 200 & 5.95 \\\hline
		\end{tabular}
		\label{reconsil}
	\end{table}

	\begin{table}[ht]
		\centering
		\small
		\caption{Test data log-probability. }
		\begin{tabular}{l|l|c}
			\hline
			Method & DIM & Log-prob\\
			\hline
			VB \cite{gan2015learning} & 200 & -136.84\\
			VB \cite{gan2015learning} & 200 - 200 & -125.60\\
			RBM \cite{cho2013enhanced} & 4000 & -107.78 \\
			DBN* \cite{salakhutdinov2008quantitative}& 200 - 200 & -120.46\\
			DBM* \cite{salakhutdinov2009deep} & 200 - 200 & -118.73\\
			\hline
			LRBN & 200 & -102.21 \\
			LRBN & 200 - 200 & -97.49 \\
			\hline
		\end{tabular}
		\label{silloglik}
	\end{table}
	
	\subsection{OCR letters dataset}

	The last experiment is performed on the OCR letters dataset, which contains 42,152 training images and 10,000 testing images of English letters. The images have the dimensionality of $16\!\times\! 8$. 
	
	The reconstruction error is reported in Table \ref{reconocr}. The proposed method shows superior performance compared to all the competing methods. The average reconstruction error on the test set is 5.95 pixels, which is at least 17 pixels better than the other methods.
	
\begin{figure}[ht]
		\centering
		\includegraphics[width=0.6\textwidth]{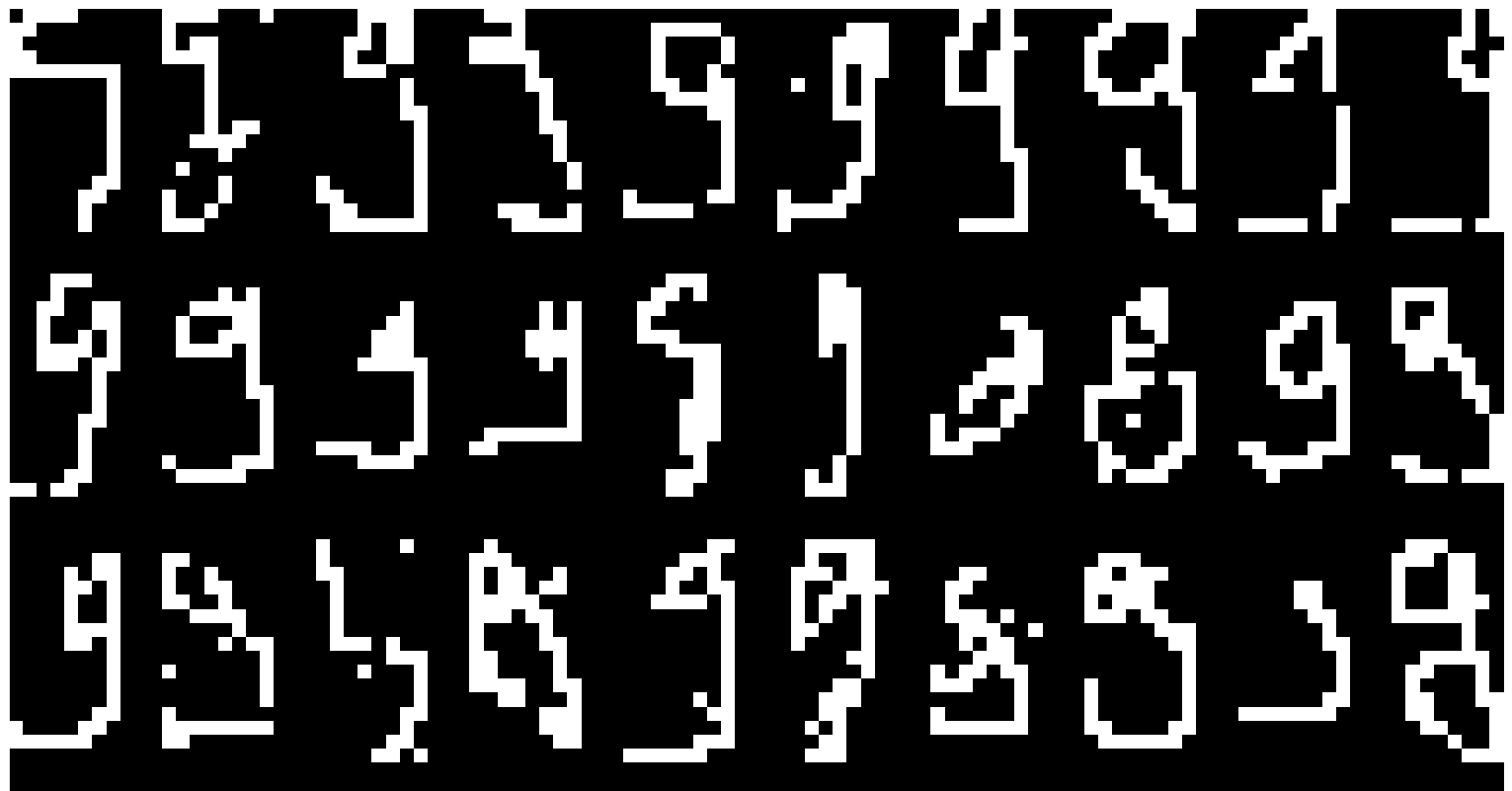}       
		\caption{Random samples from the generative model on OCR letters dataset.}
		\label{fig:letter}
	\end{figure}

	The test data log-probability is reported in Table \ref{ocrloglik}. Our model obtains a variational lower bound of -35.02, which outperforms the variational Bayes learning method, and is slightly worse than DBM \cite{salakhutdinov2010efficient}, which has 100 times more parameters. Samples from the LRBN are shown in Fig. \ref{fig:letter}. We display the samples of letter 'g'. For the same letter, the learned model is able to capture the different handwriting styles, while preserving the key information.
	
	\begin{table}[ht]
		\centering
		\small
		\caption{Average reconstruction errors of different methods on OCR letters dataset.}
		\begin{tabular}{l|l|c}
			\hline
			Method & DIM & Recon Error \\
			\hline
			NVIL* \cite{mnih2014neural} &200 - 200 & 14.79 \\
			RBM* \cite{cho2013enhanced} & 200 & 16.83\\
			DBN* \cite{salakhutdinov2008quantitative} & 200 - 200 & 12.47 \\
			DBM* \cite{salakhutdinov2009deep} & 200 - 200 & 11.14 \\\hline
			LRBN & 200 - 200 & 2.03 \\\hline
		\end{tabular}
		\label{reconocr}
	\end{table}

	\begin{table}[ht]
		\centering
		\small
		\caption{Test data log-probability on OCR letters dataset. }
		\begin{tabular}{l|l|c}
			\hline
			Method & DIM & Log-prob\\
			\hline
			VB \cite{gan2015learning} & 200 & -48.20\\
			VB \cite{gan2015learning} & 200 - 200 & -47.84\\
			DBN* \cite{salakhutdinov2008quantitative} & 200 - 200 & 40.75\\
			DBM \cite{salakhutdinov2010efficient} & 2000 - 2000 & -34.24 \\
			\hline
			LRBN & 200 & -39.48 \\
			LRBN & 200 - 200 & -35.02 \\
			\hline
		\end{tabular}
		\label{ocrloglik}
	\end{table}
	
	\section{Conclusion}
	In this work, we introduce a directed deep model based on the latent regression Bayesian network to explicitly capture the dependencies among the latent variables for data representation. We introduce an efficient inference method based on pseudo-likelihood and coordinate ascent. A hard EM learning method is proposed for efficient parameter learning. The proposed inference method solve the inference intractability, while preserving the dependencies among latent variables. We theoretically and empirically compare different models and learning methods. We point out that the latent variables in regression Bayesian network have strong dependencies, which can better explain the patterns in the input layer. Experiments on benchmark datasets shows the proposed model significantly outperforms the existing models in data reconstruction and achieves comparable performance for data representation.
	
	\bibliography{example_paper}

\begin{thebibliography}{22}
\providecommand{\natexlab}[1]{#1}
\providecommand{\url}[1]{\texttt{#1}}
\expandafter\ifx\csname urlstyle\endcsname\relax
  \providecommand{\doi}[1]{doi: #1}\else
  \providecommand{\doi}{doi: \begingroup \urlstyle{rm}\Url}\fi

\bibitem[Bengio et~al.(2014)Bengio, Yao, and Cho]{bengio2013bounding}
Yoshua Bengio, Li~Yao, and Kyunghyun Cho.
\newblock Bounding the test log-likelihood of generative models.
\newblock In \emph{International Conference on Learning Representations
  (Conference Track)}, 2014.

\bibitem[Besag(1986)]{besag1986statistical}
Julian Besag.
\newblock On the statistical analysis of dirty pictures.
\newblock \emph{Journal of the Royal Statistical Society. Series B
  (Methodological)}, pages 259--302, 1986.

\bibitem[Blei et~al.(2003)Blei, Ng, and Jordan]{blei2003latent}
David~M Blei, Andrew~Y Ng, and Michael~I Jordan.
\newblock Latent dirichlet allocation.
\newblock \emph{the Journal of machine Learning research}, 3:\penalty0
  993--1022, 2003.

\bibitem[Cho et~al.(2013)Cho, Raiko, and Ilin]{cho2013enhanced}
K~Cho, Tapani Raiko, and Alexander Ilin.
\newblock Enhanced gradient for training restricted boltzmann machines.
\newblock \emph{Neural computation}, 25\penalty0 (3):\penalty0 805--831, 2013.

\bibitem[Gan et~al.(2015)Gan, Henao, Carlson, and Carin]{gan2015learning}
Zhe Gan, Ricardo Henao, David Carlson, and Lawrence Carin.
\newblock Learning deep sigmoid belief networks with data augmentation.
\newblock \emph{International Conference on Artificial Intelligence and
  Statistics}, 2015.

\bibitem[Ghahramani and Hinton(1996)]{ghahramani1996algorithm}
Zoubin Ghahramani and Geoffrey~E Hinton.
\newblock The em algorithm for mixtures of factor analyzers.
\newblock Technical report, Technical Report CRG-TR-96-1, University of
  Toronto, 1996.

\bibitem[Gregor et~al.(2014)Gregor, Mnih, and Wierstra]{gregor2013deep}
Karol Gregor, Andriy Mnih, and Daan Wierstra.
\newblock Deep autoregressive networks.
\newblock \emph{In Proceedings of the 31st International Conference on Machine
  Learning}, 2014.

\bibitem[Hinton et~al.(2006)Hinton, Osindero, and Teh]{hinton2006fast}
Geoffrey Hinton, Simon Osindero, and Yee-Whye Teh.
\newblock A fast learning algorithm for deep belief nets.
\newblock \emph{Neural computation}, 18\penalty0 (7):\penalty0 1527--1554,
  2006.

\bibitem[Hinton et~al.(1995)Hinton, Dayan, Frey, and Neal]{hinton1995wake}
Geoffrey~E Hinton, Peter Dayan, Brendan~J Frey, and Radford~M Neal.
\newblock The" wake-sleep" algorithm for unsupervised neural networks.
\newblock \emph{Science}, 268\penalty0 (5214):\penalty0 1158--1161, 1995.

\bibitem[Hofmann(1999{\natexlab{a}})]{hofmann1999probabilistic1}
Thomas Hofmann.
\newblock Probabilistic latent semantic indexing.
\newblock In \emph{Proceedings of the 22nd annual international ACM SIGIR
  conference on Research and development in information retrieval}, pages
  50--57. ACM, 1999{\natexlab{a}}.

\bibitem[Hofmann(1999{\natexlab{b}})]{hofmann1999probabilistic2}
Thomas Hofmann.
\newblock Probabilistic latent semantic analysis.
\newblock In \emph{Proceedings of the Fifteenth conference on Uncertainty in
  artificial intelligence}, pages 289--296. Morgan Kaufmann Publishers Inc.,
  1999{\natexlab{b}}.

\bibitem[Kingma and Welling(2014)]{kingma2013auto}
Diederik~P Kingma and Max Welling.
\newblock Auto-encoding variational bayes.
\newblock \emph{In Proceedings of the International Conference on Learning
  Representations (ICLR)}, 2014.

\bibitem[Mnih and Gregor(2014)]{mnih2014neural}
Andriy Mnih and Karol Gregor.
\newblock Neural variational inference and learning in belief networks.
\newblock \emph{In Proceedings of the 31st International Conference on Machine
  Learning}, 2014.

\bibitem[Neal(1992)]{neal1992connectionist}
Radford~M Neal.
\newblock Connectionist learning of belief networks.
\newblock \emph{Artificial intelligence}, 56\penalty0 (1):\penalty0 71--113,
  1992.

\bibitem[Patel et~al.(2015)Patel, Nguyen, and Baraniuk]{patel2015probabilistic}
Ankit~B Patel, Tan Nguyen, and Richard~G Baraniuk.
\newblock A probabilistic theory of deep learning.
\newblock \emph{arXiv preprint arXiv:1504.00641}, 2015.

\bibitem[Rezende et~al.(2014)Rezende, Mohamed, and
  Wierstra]{rezende2014stochastic}
Danilo~J Rezende, Shakir Mohamed, and Daan Wierstra.
\newblock Stochastic backpropagation and approximate inference in deep
  generative models.
\newblock In \emph{Proceedings of the 31st International Conference on Machine
  Learning (ICML-14)}, pages 1278--1286, 2014.

\bibitem[Salakhutdinov and Hinton(2009)]{salakhutdinov2009deep}
Ruslan Salakhutdinov and Geoffrey~E Hinton.
\newblock Deep boltzmann machines.
\newblock In \emph{International Conference on Artificial Intelligence and
  Statistics}, pages 448--455, 2009.

\bibitem[Salakhutdinov and Larochelle(2010)]{salakhutdinov2010efficient}
Ruslan Salakhutdinov and Hugo Larochelle.
\newblock Efficient learning of deep boltzmann machines.
\newblock In \emph{International Conference on Artificial Intelligence and
  Statistics}, pages 693--700, 2010.

\bibitem[Salakhutdinov and Murray(2008)]{salakhutdinov2008quantitative}
Ruslan Salakhutdinov and Iain Murray.
\newblock On the quantitative analysis of deep belief networks.
\newblock In \emph{Proceedings of the 25th international conference on Machine
  learning}, pages 872--879. ACM, 2008.

\bibitem[Saul et~al.(1996)Saul, Jaakkola, and Jordan]{saul1996mean}
Lawrence~K Saul, Tommi Jaakkola, and Michael~I Jordan.
\newblock Mean field theory for sigmoid belief networks.
\newblock \emph{Journal of Artificial Intelligence Research}, 4\penalty0
  (61):\penalty0 76, 1996.

\bibitem[Tang et~al.(2012)Tang, Hinton, and Salakhutdinov]{tang2012deep}
Yichuan Tang, Geoffrey~E Hinton, and Ruslan Salakhutdinov.
\newblock Deep mixtures of factor analysers.
\newblock In \emph{Proceedings of the 29th International Conference on Machine
  Learning (ICML-12)}, pages 505--512, 2012.

\bibitem[van~den Oord and Schrauwen(2014)]{van2014factoring}
Aaron van~den Oord and Benjamin Schrauwen.
\newblock Factoring variations in natural images with deep gaussian mixture
  models.
\newblock In \emph{Advances in Neural Information Processing Systems}, pages
  3518--3526, 2014.

\end{thebibliography}
	\bibliographystyle{plainnat}

\end{document}